\documentclass{article}

\PassOptionsToPackage{numbers, compress}{natbib}
\usepackage[final]{nips}

\usepackage[utf8]{inputenc} 
\usepackage[T1]{fontenc}    
\usepackage{hyperref}       
\usepackage{url}            
\usepackage{booktabs}       
\usepackage{amsfonts}       
\usepackage{nicefrac}       
\usepackage{microtype}      

\usepackage{epsfig}
\usepackage{graphicx}
\usepackage{amsmath}
\usepackage{amssymb}
\usepackage{algorithm,algorithmic}
\usepackage{subfigure} 
\usepackage{multirow}
\usepackage{bm}
\usepackage{hyphenat}

\title{Distributed Deep Transfer Learning \\ by Basic Probability Assignment}

\author{
  Arash shahriari \\
  Research School of Engineering \\
  Australian National University \\
  Canberra, Australia \\
  \texttt{arash.shahriari@anu.edu.au} \\
}

\begin{document}

\maketitle

\begin{abstract}
\nohyphens{Transfer learning is a popular practice in deep neural networks, but fine-tuning of large number of parameters is a hard task due to the complex wiring of neurons between splitting layers and imbalance distributions of data in pretrained and transferred domains. The reconstruction of the original wiring for the target domain is a heavy burden due to the size of interconnections across neurons. We propose a distributed scheme that tunes the convolutional filters individually while backpropagates them jointly by means of basic probability assignment. Some of the most recent advances in evidence theory show that in a vast variety of the imbalanced regimes, optimizing of some proper objective functions derived from contingency matrices prevents biases towards high-prior class distributions. Therefore, the original filters get gradually transferred based on individual contributions to overall performance of the target domain. This largely reduces the expected complexity of transfer learning whilst highly improves precision. Our experiments on standard benchmarks and scenarios confirm the consistent improvement of our distributed deep transfer learning strategy.}
\end{abstract}

\section{Introduction}

Transfer learning is a popular practice in deep neural networks. Fine-tuning of a large number of parameters, is difficult due to the complex wiring of neurons between splitting layers. The imbalanced distributions of data for primary and target domains, also adds up to the hurdle of the problem. The reconstruction of the primary wiring for the target network is a heavy burden, considering the size of interconnections across neurons. 

For supervised learning, many classification algorithms assume the same distributions for training and test data. In order to change this distribution, the statistical models must be rebuilt. This is not always practical, due to the difficulty of recollecting the training data or complexity of the learning process. One of the solutions is transfer learning, which transfers the classification knowledge into a new domain~(\cite{pan2010survey}). The aim is to learn highly generalized models for either domains with different probability distributions, or domains without labelled data~(\cite{wang2014flexible},~\cite{zhang2013domain}). Here, the main challenge is to reduce the shifts in data distributions between the domains, by using algorithms that minimize their discrimination. It is worth mentioning that, this cannot resolve domain-specific variations~(\cite{long2016deep}).

Transfer learning has also proven to be highly beneficial at boosting the overall performance of deep neural networks. Deep learning practices usually require a huge amount of labelled data to learn powerful models. The transfer learning enables adaptation to a different source with a small number of training samples. 

\newpage

On the other hand, deep neural networks practically learn intermediate features. They can provide better transfer learning, because some of them generalize well, among various domains of knowledge~(\cite{glorot2011domain}). These transferable features generally underlie several probability distributions~(\cite{oquab2014learning}), to reduce the cross-domain discrepancy~(\cite{yosinski2014transferable}). The common observation among several deep architectures, is that, features learned in the bottom layers are not that specific, but transiting towards top layers, tailors them to the target dataset or task. 

A recent study of the generality or specificity of deep layers for transfer learning, show two difficulties which may affect the transfer of deep features~(\cite{yosinski2014transferable}). First, top layers get quite specialized to their primary tasks and second, some optimization difficulties arise due to the splitting of the network between adapted layers. In spite of these negative effects, other studies have confirmed that transferred features, not only perform better than random representation, but also provide better initialization. 

This paper proposes a distributed backpropagation scheme in which, the convolutional filters are fine-tuned individually, but are backpropagated, all at once. This is done by means of Basic Probability Assignment~(\cite{sentz2002combination}) of evidence theory. Therefore, the primary filters are gradually transferred, based on, their contributions to classification performance of the target domain. This approach largely reduces the complexity of transfer learning, whilst improving precision. The experimental results on standard benchmarks and various scenarios, confirm the consistent improvement of the distributed backpropagation strategy for the transfer learning.

\section{Method}
\label{sec:method}

A novel framework for distributed backpropagation in deep convolutional networks is introduced, that alleviates the burden of splitting a network through the middle of fragile layers. The intuition is that, this difficulty relates to the complexity of deep architecture and the imbalance data distribution of the primary and target domains. 

On one hand, the splitting of layers, results in optimization difficulty, because there is a high complexity in the interconnections between neurons of adapted layers. To address this, the convolutional filters are fine-tuned individually. This reduces the complexity of non-convex optimization for the transfer learning problem. On the other hand, the imbalance problem arises form different distributions of data, in the primary and target domains. This issue can be handled by cost-sensitive, imbalanced learning methods. Since the power of deep neural models, comes from mutual optimization of all parameters, the above fine-tuned convolutional filters are joined by a cost-sensitive backpropagation scheme.

The emergence of new cost-sensitive methods for imbalanced data~(\cite{elkan2001foundations}), enables the misclassification costs to be embedded in the form of a cost matrix. Meaningful information is then able to be distributed to the learning process. The error, based on the misclassification costs, is measured for each class, to form a confusion matrix. This matrix is the most informative contingency table in imbalanced  learning problems, because it gives the success rate of a classifier in a special class, and the failure rate on distinguishing that class from other classes. The confusion matrix has proven to be a great regularizer; smoothing the accuracy among imbalanced data and giving more importance to minority distributions~(\cite{ralaivola2012confusion}). 

Determination of a probabilistic distribution from the confusion matrix, is highly effective at producing a probability assignment, which contributes to the imbalanced problems. This can be either constructed from recognition, substitution and rejection rates~(\cite{xu1992methods}) or both precision and recall rates~(\cite{deng2016improved}). The key point is to harvest maximum possible prior knowledge from the confusion matrix, to overcome an imbalanced challenge. The experiments confirm the advantage of the proposed distributed backpropagation for transfer learning in deep convolutional neural networks. 

\newpage

\section{Formulation}
\label{sec:formulation}

It is a general practice in transfer learning to include training of a primary deep neural network on a dataset, followed by fine-tuning of learned features for another dataset, on a new target network~(\cite{bengio2012deep}). The generality of selected features for both of the primary and target domains, is critical to the success of transfer learning. 

\begin{figure}[!t]
\begin{center}
\includegraphics[width=0.8\textwidth]{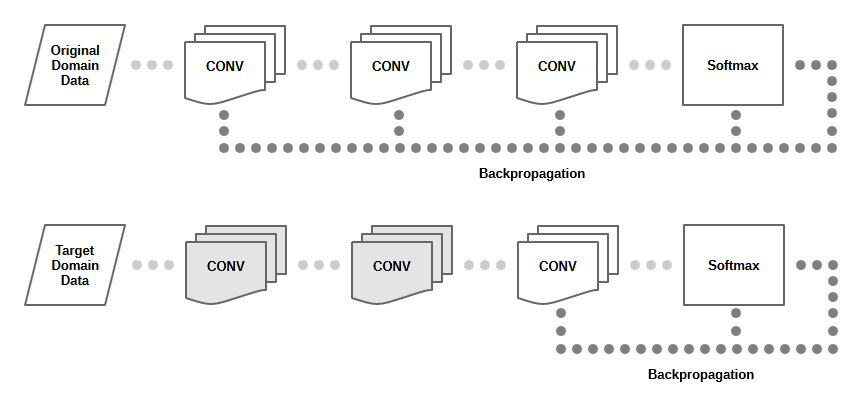}
\end{center}
\caption[Standard backpropagation for transfer learning.]{Standard backpropagation for transfer learning. The parameters of the target network can be either updated, in all layers or be frozen in the bottom layers, during backpropagation. Overfitting is highly probable for a small target dataset on a large primary network, due to the number of parameters. In contrast, underfitting is possible for a large target dataset on a small primary network, because of imbalanced data distributions. To fine-tune the pretrained network (primary) for the transferred network (target), backpropagation is applied to the top layers of the deep network, fed by the target data, while bottom layers remain fixed.}
\label{fig:object:transfer:standard}
\end{figure}

For implementation, the primary network is trained and its bottom layers copied, to form the target network. The top layers of the target network, are initialized randomly and are trained by the target data. It is possible to employ backpropagation for all layers and fine-tune their parameters for the target task, or freeze the copied primary layers, and only update the top target layers. This is usually decided by the size of the target dataset and number of parameters in the primary layers. Fine-tuning of large networks for small datasets, leads to overfitting, but for small networks and large datasets, improves the performance ~(\cite{sermanet2013overfeat}). A diagram of standard backpropagation for transfer learning is presented in Figure~\ref{fig:object:transfer:standard}.

To handle the overfitting issues, a distributed backpropagation paradigm is proposed. First, the large number of fine-tuning parameters are divided, so as, to conquer the complexity of the primary non-convex optimization. This is implemented by breaking of the $l$-layer network, of depth $d$, to the $d$ distributed $l$-layer single-filter networks, of depth one. This leads to a decay rate of $1/d$ in the number of parameters, needing to be fine-tuned in each single-filter networks. Second, this distributed architecture is fed with the target data. As a result, each of the single-filter networks, generates a contingency matrix of its Softmax classifier for the classification, recognition, or regression task.

\begin{figure}[!t]
\begin{center}
\includegraphics[width=1.0\textwidth]{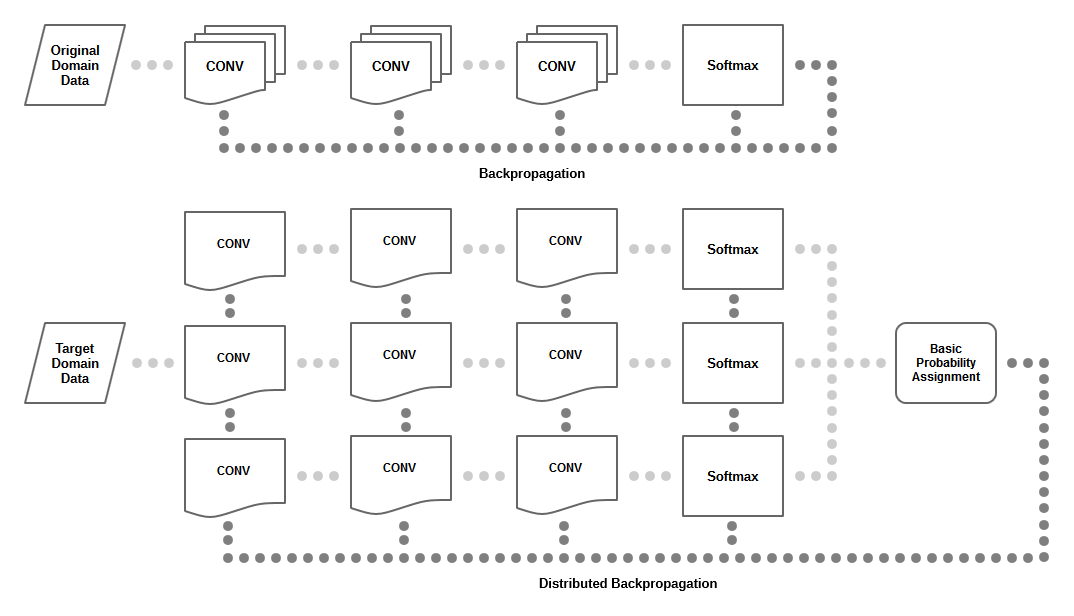}
\end{center}
\caption[Distributed backpropagation for transfer learning.]{Distributed backpropagation for transfer learning. The primary network, pretrained on the primary domain, is arranged in several single-filter networks, each of which fine-tunes a single convolutional filter. For backpropagation, BPA is employed to determine the contribution of each single-filter networks. The computed costs are multiplied by the error gradients to backpropagate through the distributed single-filter networks, which are now fed by the target data. The distributed backpropagation controls the rate of change in weights and biases by a cost-sensitive formulation. As a result, the primary convolutional filters contribute significantly to the performance of the target domain, generating larger steps for gradient descent optimization. In contrast, the primary filters with small probability assignments, stop iterative optimization by causing infinitesimal steps. This leads to a powerful transferred model, which is better tailored to the target data distribution.}
\label{fig:object:transfer:distributed}
\end{figure}

Third, BPA is employed to find out the contribution of each single-filter networks, to the specific learning task, in the target domain. Finally, the calculated probability assignments are normalized, to use as the costs of backpropagation, in each of the single-filter networks. In other words, the parameters are updated by the multiplication of error gradients, and the cost of each distributed single-filter networks~(Figure~\ref{fig:object:transfer:distributed}).

The single-filter architectures are initialized by parameters, learned from the primary domain, meanwhile, they are optimized by gradient descent through backpropagation on the target domain. It differs from ensemble learning in that, single-filter networks are not re-weighted, but their parameters are updated to cope with the target domain.

\subsection{Basic Probability Assignment}
\label{sec:assignment}

A confusion matrix is generally represented as class-based predictions against actual labels, in the form of a square matrix. Inspired by Dempster-Schafer theory~(\cite{sentz2002combination}), construction of BPA gives a vector, which is independent of the number of samples in each classes and sums up to one, for each labels. BPA provides the ability to reflect the different contributions of a classifier, or combine the outcomes of multiple weak classifiers. A raw, two-dimensional confusion matrix, indexed by the predicted classes and actual labels, provides some common measures of classification performance. Some general measures are accuracy (the proportion of the total number of predictions that are correct), precision (a measure of the accuracy, provided that a specific class has been predicted), recall (a measure of the ability of a prediction model to select instances of a certain class from a dataset), and F-score (the harmonic mean of precision and recall)~(\cite{sammut2011encyclopedia}).

Suppose that a set of training samples $\mathcal{X}=\{X_1,\dots,X_{|\mathcal{X}|}\}$ from $\mathcal{C}=\{C_1,\dots,C_{|\mathcal{C}|}\}$ different classes, are assigned to a label set $\mathcal{L}=\{L_1,\dots,L_{|\mathcal{L}|}\}$, using a classifier $\phi$. Each element $n_{ij}$ of the confusion matrix $\bm{\Upsilon}_{\phi}$ is considered as the number of samples belonging to class $C_i$, which assigned to label $L_j$. The recall ($r_{ij}$) and precision ($s_{ij}$) ratios for all $i\in\{1\dots{|\mathcal{C}|}\}$ and $j\in\{1\dots{|\mathcal{L}|}\}$, can be defined as follows~(\cite{deng2016improved}),

\begin{eqnarray}
r_{ij} &=& \frac{n_{ij}}{\sum_{i=1}^{|\mathcal{C}|}{n_{ij}}}\nonumber\\
s_{ij} &=& \frac{n_{ij}}{\sum_{j=1}^{|\mathcal{L}|}{n_{ij}}}
\label{eq:object:14}
\end{eqnarray}

It can be seen that, the recall ratio is summed over the predicted classes (rows), whilst the precision ratio is accumulated by the actual labels (columns) of the confusion matrix $\bm{\Upsilon}_{\phi}$. The probability elements of recall ($R_{i}$) and precision ($S_{i}$) for each individual class $C_i$ are, 

\begin{eqnarray}
R_{i}&=&\frac{r_{ii}}{\sum_{j=1}^{|\mathcal{C}|}{r_{jj}}}\nonumber\\
S_{i}&=&\frac{s_{ii}}{\sum_{j=1}^{|\mathcal{L}|}{s_{jj}}} 
\label{eq:object:15}
\end{eqnarray}

\begin{algorithm}[!t]
\caption{Basic Probability Assignment}
\label{alg:object:transfer:bpa}
\begin{algorithmic}
\STATE
\STATE {\bfseries Input:} training set $\mathcal{X}$, classifier $\phi$
\STATE {\bfseries Output:} basic probability assignment $\bm{\Theta}_{\phi}$
\STATE
\STATE 1: Compute recall and precision~(Equation~\ref{eq:object:14})
\STATE 2: Calculate recall and precision assignments (Equation~\ref{eq:object:15})
\STATE 3: Apply Dempster-Schafer rule of combination (Equation~\ref{eq:object:16})
\STATE
\end{algorithmic}
\end{algorithm}

These elements are synthesized to form the final probability assignments by Dempster-Schafer rule of combination~(\cite{sentz2002combination}), representing the recognition ability of classifier $\phi$ to each of the classes of set $\mathcal{C}$ as,

\begin{equation}
M_i = R_i\oplus{S_i} = \frac{R_i{S_i}}{1-\sum_{i=1}^{|\mathcal{C}|}R_i{S_i}}
\label{eq:object:16}
\end{equation}

\noindent where the operator $\oplus$ is an orthogonal sum. The overall contribution of the classifier $\phi$ can be presented as a probability assignment vector,

\begin{equation}
\bm{\Theta}_{\phi}=\{M_i\;|\;\forall\;i\in[1,|\mathcal{C}|]\}
\label{eq:object:17}
\end{equation}

It is worth mentioning that $\bm{\Theta}_{\phi}$ should be computed by the training set, because it is assumed that, there is no actual label set $\mathcal{L}$ at the test time.

\subsection{Distributed Backpropagation}
\label{sec:transfer}

Suppose that $\Phi=\{\phi_1,\dots,\phi_{|\Phi|}\}$ is a set of Softmax loss functions of the single-filter networks, presented in Figure~\ref{fig:object:transfer:distributed}. To apply the distributed backpropagation, Algorithm~\ref{alg:object:transfer:bpa} should be followed for each of the classifiers. The result is a set of normalized probability assignments as follows,

\begin{equation}
\bm{\Gamma}_{\Phi}=\{\bm{\Gamma}_{\phi_k}=\lVert{\bm\Theta_{\phi_k}}\rVert_2\;|\;\forall\;k\in[1,|\Phi|]\}
\label{eq:object:18}
\end{equation}

\begin{algorithm}[!t]
\caption{Distributed Backpropagation for Transfer Learning}
\label{alg:object:transfer:dtl}
\begin{algorithmic}
\STATE
\STATE {\bfseries Input:} primary network, target data 
\STATE {\bfseries Output:} fine-tuned target network
\STATE
\STATE 1: Compute $\bm{\Gamma}_{\Phi}$~(Equation~~\ref{eq:object:18}) by using Algorithm~\ref{alg:object:transfer:bpa}
\STATE
\FOR{$k=1$ {\bfseries to} $|\Phi|$}
\FORALL{layers}
\STATE 2: Compute feedforward activations~(Equation~~\ref{eq:object:19})
\STATE 3: Calculate backpropagation errors~(Equation~~\ref{eq:object:21})
\STATE 4: Update weights and biases~(Equation~~\ref{eq:object:22})
\ENDFOR
\ENDFOR
\STATE
\end{algorithmic}
\end{algorithm}

It is known that in each layer $l$ of the $k$th single-filter network, the feedforward propagation is calculated as,

\begin{equation}
a^{(l)} = \sigma(w^{(l)}{a^{(l-1)}}+b^{(l)})
\label{eq:object:19}
\end{equation}

\noindent which $w$ and $b$ are weights and biases, $a$ is an activation and $\sigma$ is a rectification function. Considering $\phi$ as the cost function of the $k$th network, the output error holds,

\begin{equation}
\delta^{(l)} = \bigtriangledown_a\phi\odot\sigma^{'}(w^{(l)}{a^{(l-1)}}+b^{(l)})
\label{eq:object:20}
\end{equation}

\noindent and the backpropagation error can be stated as,

\begin{equation}
\delta^{(l)} = ((w^{(l+1)})^T)\delta^{(l+1)}\odot\sigma^{'}(w^{(l)}{a^{(l-1)}}+b^{(l)})
\label{eq:object:21}
\end{equation}

For the sake of gradient descent, the weights and biases are updated via,

\begin{eqnarray}
w^{(1)}&\longrightarrow&w^{(1)}-\eta\;\Gamma_{\phi}\sum\delta^{(l+1)}(a^{(l-1)})^T\nonumber\\
b^{(1)}&\longrightarrow&b^{(1)}-\eta\;\Gamma_{\phi}\sum\delta^{(l+1)}
\label{eq:object:22}
\end{eqnarray}

It can be seen that, greater $\bm\Gamma_{\phi}$ of the $k$th single-filter network, makes larger steps to update the weights and biases of the target domain, during distributed backpropagation, compared to the standard backpropagation, with a fix $\eta$ in the primary domain. This helps to only update primary filters, which largely affect the target domain and also properly connect the distributed single-filter networks, during the fine-tuning process. This also implies that, in spite of forward-backward propagation in the target domain, the overall contribution of all convolutional filters, is taken into account. Since the united backpropagation is involved in every iterations of training, the single-filter networks are trained together, which is different from ensemble learning. Algorithm~\ref{alg:object:transfer:dtl} wraps up the proposed strategy.

It is assumed that the number of classes and assigned labels are equal ($|\mathcal{C}|=|\mathcal{L}|$), although the merging of different classes is a common practice, particularly in visual classification (for example, vertical vs horizontal categories). The benefit lies in the fact that the bottom layers of deep convolutional architectures, contribute to detecting first and second order features, that are usually of specific directions, rather than specific distinguishing patterns of the objects. This leads to a powerful hierarchical feature learning, in the case $|\mathcal{C}|\ll|\mathcal{L}|$. In contrast, some classes can be divided into various subcategories, although they all get the same initial labels, and hence this holds $|\mathcal{C}|\gg|\mathcal{L}|$ to take the advantage of more general features in the top layers. The proposed distributed backpropagation does not merge or divide the primary labels of the datasets under study, although it seems that, this is able to boost the performance of transfer learning, for both of the merging or dividing cases.

\section{Experiments}
\label{experiments}

Two different scenarios are considered to evaluate the performance of distributed backpropagation for transfer learning. In the first scenario, the performance of fine-tuning for pairs of datasets, with either close data distributions or number of classes, are observed. These are MNIST \& SVHN and CIFAR-10 \& CIFAR-100 pairs, as primary \& target domains, and the performance of transfer learning, are reported in the form of training-test errors. For the second scenario, transfer learning is applied for pairs of datasets, with far data-class distributions, which are MNIST \& CIFAR-10 and SVHN \& CIFAR-100 pairs. In these experiments, the datasets are arranged to examine the effect of dissimilar distributions, rather than overfitting.

\begin{table}[!t]
\begin{center}
\begin{tabular}{|c||c|c|}
\hline
\multicolumn{1}{|c||}{\multirow{2}{*}{Dataset}} & \multicolumn{2}{c|}{Baseline}\\
\cline{2-3}
\multicolumn{1}{|c||}{} & \multicolumn{1}{c|}{Train~(\%)} & \multicolumn{1}{c|}{Test~(\%)}\\
\hline\hline
\multicolumn{1}{|c||}{MNIST} & 0.04 & 0.55\\
\multicolumn{1}{|c||}{SVHN} & 0.13 & 3.81\\
\multicolumn{1}{|c||}{CIFAR-10} & 0.01 & 19.40\\
\multicolumn{1}{|c||}{CIFAR-100} & 0.17 & 50.90\\
\hline
\end{tabular}
\end{center}
\caption[Baselines of classification errors on standard benchmarks.]{Baselines of classification errors on standard benchmarks. The results are produced by deep learning of convolutional neural networks, on the datasets under examination.}
\label{tb:baseline}
\end{table}

\newpage

As a practical example, suppose that the aim is to transfer MNIST as the primary domain, to CIFAR as the target domain. To initialize, CIFAR data is presented to the MNIST pretrained network, filter by filter, for each of the 20 convolutional filters of the MNIST network. Then, the outputs of Softmax layers are passed to BPA module, and the error gradients are multiplied by the specific cost of each filters. These new gradients backpropagate to all 20 single-filter networks, to update their weights and biases. After some iterations, the MNIST network is transferred to the CIFAR network, through the distributed backpropagation scheme, running on 20 fine-tuned, single-filter networks. The baselines of training and test errors on the experimental datasets, are reported in Table~\ref{tb:baseline}. For ease of implementation and fast replication of the results, the deep learning library provided by the Oxford Visual Geometry Group~(\cite{vedaldi08vlfeat}), is deployed.

\subsection{Transfer Learning on Fairly Balanced Domains}
\label{exp1}

In this scenario, two pairs of datasets are targeted, which contain similar data distributions and perform the same recognition tasks. The results are reported for standard vs distributed backpropagations, in Table~\ref{tb:exp1}. The standard backpropagation trains the primary network and fine-tunes the top layers for the target network~(\cite{bengio2012deep}). The proposed distributed backpropagation employs BPA, for the cost-sensitive transfer learning.

It can be seen that, the results for the standard backpropagation follow the argument on size of networks and number of model parameters~(\cite{sermanet2013overfeat}). MNIST does a poor job on transferring to SVHN, due to the overfitting of SVHN over MNIST network. In contrast, SVHN performs well as the primary network to transfer MNIST as the target domain.

On the other hand, transferring to SVHN domain from MNIST, does not result in overfitting, when the distributed backpropagation is employed. In both settings of the primary and target domains, the distributed strategy outperforms the standard backpropagation. 

The experiments based on the CIFAR pair, raise interesting results, because both datasets have the same number of samples, but completely different distributions among the classes. In practice, CIFAR-100 includes all the classes of CIFAR-10, but CIFAR-10 is not aware of several classes in CIFAR-100. It can be seen that, CIFAR-10 transfers well to CIFAR-100. This cannot outperform the baseline performance, although the target network (CIFAR-100) is not overfitted. All in all, the performance of the distributed backpropagation for transfer learning is better than the standard scheme and also, outperforms the baselines of the benchmarks. 

\subsection{Transfer Learning on Highly Imbalanced Domains}
\label{exp2}

This scenario pairs the datasets, such that, the similarity of their data distributions, and the number of classes get minimized. They are also initially trained for different tasks, \textit{i.e.} number classification vs object recognition. For implementation, MNIST gray channel is repeated three times to make a RGB-like colour representation for transferring to CIFAR network. For the CIFAR, the RGB channels are converted into grayscale to transfer on the MNIST network.

From Table~\ref{tb:exp2}, it is obvious that the distributed backpropagation outperforms all the standard results. For the first setup, CIFAR-10 performs better at transfer learning than MNSIT, although the number of classes are the same. 

\begin{table}[!t]
\begin{center}
\begin{tabular}{|c|c||c|c||c|c|}
\hline
\multicolumn{1}{|c|}{\multirow{2}{*}{primary}} & \multicolumn{1}{c||}{\multirow{2}{*}{Target}} & \multicolumn{2}{c||}{Standard} & \multicolumn{2}{c|}{Distributed}\\
\cline{3-6}
\multicolumn{1}{|c|}{} & \multicolumn{1}{c||}{} & \multicolumn{1}{c|}{Train~(\%)} & \multicolumn{1}{c||}{Test~(\%)} & \multicolumn{1}{c|}{Train~(\%)} & \multicolumn{1}{c|}{Test~(\%)}\\
\hline\hline
\multicolumn{1}{|c|}{MNIST} & \multicolumn{1}{c||}{SVHN} & \textbf{0.01} & 29.57 & 0.24 & \textbf{5.18}\\
\multicolumn{1}{|c|}{SVHN} & \multicolumn{1}{c||}{MNIST} & \textbf{0.35} & 1.04 & 0.16 & \textbf{0.46}\\
\multicolumn{1}{|c|}{CIFAR-10} & \multicolumn{1}{c||}{CIFAR-100} & 0.53 & 68.44 & \textbf{0.29} & \textbf{54.32}\\
\multicolumn{1}{|c|}{CIFAR-100} & \multicolumn{1}{c||}{CIFAR-10} & 0.11 & 24.08 & \textbf{0.05} & \textbf{18.24}\\
\hline
\end{tabular}
\end{center}
\caption[Classification errors of standard vs distributed backpropagations, for transfer learning on fairly balanced domains.]{Classification errors of standard vs distributed backpropagations, for transfer learning on fairly balanced domains. Bold values correspond to better performances, which are the smallest training and test errors on each experiment.}
\label{tb:exp1}
\end{table}

\begin{table}[!t]
\begin{center}
\begin{tabular}{|c|c||c|c||c|c|}
\hline
\multicolumn{1}{|c|}{\multirow{2}{*}{primary}} & \multicolumn{1}{c||}{\multirow{2}{*}{Target}} & \multicolumn{2}{c||}{Standard} & \multicolumn{2}{c|}{Distributed}\\
\cline{3-6}
\multicolumn{1}{|c|}{} & \multicolumn{1}{c||}{} & \multicolumn{1}{c|}{Train~(\%)} & \multicolumn{1}{c||}{Test~(\%)} & \multicolumn{1}{c|}{Train~(\%)} & \multicolumn{1}{c|}{Test~(\%)}\\
\hline\hline
\multicolumn{1}{|c|}{MNIST} & \multicolumn{1}{c||}{CIFAR-10} & 0.43 & 28.92 & \textbf{0.25} & \textbf{20.85}\\
\multicolumn{1}{|c|}{CIFAR-10} & \multicolumn{1}{c||}{MNIST} & 0.44 & 2.37 & \textbf{0.23} & \textbf{0.95}\\
\multicolumn{1}{|c|}{SVHN} & \multicolumn{1}{c||}{CIFAR-100} & 0.71 & 89.31 & \textbf{0.46} & \textbf{61.10}\\
\multicolumn{1}{|c|}{CIFAR-100} & \multicolumn{1}{c||}{SVHN} & \textbf{0.01} & 12.18 & 0.28 & \textbf{7.25}\\
\hline
\end{tabular}
\end{center}
\caption[Classification errors of standard vs distributed backpropagations, for transfer learning on highly imbalanced domains.]{Classification errors of standard vs distributed backpropagations, for transfer learning on highly imbalanced domains. Bold values correspond to better performances, which are the smallest training and test errors on each experiment.}
\label{tb:exp2}
\end{table}

\newpage

It seems that, CIFAR-10 provides better generalization due to higher diversity among its classes. Here, the distributed backpropagation performs better than the standard process and, targeting of MNIST from CIFAR-10 network, results in a performance that is similar to the baseline outcomes on MNIST in Table~\ref{tb:baseline}. The second setup leads to the overfitting of SVHN over CIFAR-100 network, as a result of the large number of samples. The other outcome is the poor performance of transferring CIFAR-100 to SVHN network. This is the result of different contents of primary and target datasets. 

The observations show that fine-tuning on the training set, while calculating BPA on the validation set, result in better generalization of the transferred model. On the other hand, computing of BPA on training plus validation sets, gives a higher performance. This is due to the vastly different number of classes in the primary-target domains. Since BPA is employed to address the imbalance distribution problem, it better captures the distribution of data, by adjoining both training and validation sets. This is especially true, when fewer classes of the primary domain, are transferred to the larger number of classes in the target domain.

\section{Conclusion}
\label{conclusion}

We introduce a novel transfer learning for deep convolutional networks that tackles the optimization complexity of a highly non-convex objective by breaking it to several distributed fine-tuning operations which backpropagate jointly. This also resolves the imbalance learning regime for the original and target domains by using the basic probability assignment of evidence theory across several unit-depth single-filter networks. By distributed backpropagation, the overall performance shows considerable improvement over standard transfer learning scheme. We conduct several experiments on publicly available datasets and report the performance as training and test errors. The results confirm the advantage of our distributed strategy.

\newpage

\bibliographystyle{nips}
\bibliography{nips}

\end{document}